\documentclass[conference]{IEEEtran}
\IEEEoverridecommandlockouts
\usepackage{cite}
\usepackage{amsmath,amssymb,amsfonts}
\usepackage{graphicx, caption}
\usepackage{textcomp}
\usepackage{xcolor}

\usepackage[
  colorlinks,
  citecolor=red
]{hyperref}

\newtheorem{lemma}{Lemma}
\newtheorem{proof}{Proof}

\usepackage{algorithm}  
\usepackage{algorithmicx}  
\usepackage{algpseudocode}
\usepackage{wrapfig}
\usepackage{multirow} 
\usepackage{makecell} 
\usepackage{tabularx}
\usepackage{booktabs}

\def\BibTeX{{\rm B\kern-.05em{\sc i\kern-.025em b}\kern-.08em
    T\kern-.1667em\lower.7ex\hbox{E}\kern-.125emX}}
\begin{document}

\newcommand\blfootnote[1]{%
  \begingroup
  \renewcommand\thefootnote{}\footnote{#1}%
  \addtocounter{footnote}{-1}%
  \endgroup
}

\title{Efficient Meta-Learning for Continual Learning with Taylor Expansion Approximation\\
\thanks{
This work was supported by NSFC Tianyuan Fund for Mathematics (No. 12026606), National Key R\&D Program of China (No. 2018AAA0100300), and Beijing Academy of Artificial Intelligence(BAAI).}
}

\author{
\IEEEauthorblockN{1\textsuperscript{st} Xiaohan Zou}
\IEEEauthorblockA{
\textit{Boston University} \\
zxh@bu.edu}
\and
\IEEEauthorblockN{2\textsuperscript{nd} Tong Lin \thanks{Correspondence to Tong Lin (lintong@pku.edu.cn).}}
\IEEEauthorblockA{
\textit{Key Lab. of Machine Perception (MoE), School of AI,}\\
\textit{Center for Data Science, Peking University}\\
lintong@pku.edu.cn}
}

\maketitle

\begin{abstract}
Continual learning aims to alleviate catastrophic forgetting when handling consecutive tasks under non-stationary distributions. Gradient-based meta-learning algorithms have shown the capability to implicitly solve the transfer-interference trade-off problem between different examples. However, they still suffer from the catastrophic forgetting problem in the setting of continual learning, since the past data of previous tasks are no longer available. In this work, we propose a novel efficient meta-learning algorithm for solving the online continual learning problem, where the regularization terms and learning rates are adapted to the Taylor approximation of the parameter's importance to mitigate forgetting. The proposed method expresses the gradient of the meta-loss in closed-form and thus avoid computing second-order derivative which is computationally inhibitable. We also use Proximal Gradient Descent to further improve computational efficiency and accuracy. Experiments on diverse benchmarks show that our method achieves better or on-par performance and much higher efficiency compared to the state-of-the-art approaches.
\end{abstract}


\section{Introduction}

Catastrophic forgetting \cite{french1999catastrophic}, \cite{mcclelland1995there} poses a major challenge to artificial intelligence systems: when switching to a new task, the system performance may degrade on the previously trained tasks. Continual learning is proposed to address this challenge, which requires models to be stable enough to prevent forgetting while being flexible to acquire new knowledge.

To alleviate catastrophic forgetting, several categories of continual learning methods have been proposed to penalize neural networks with regularization approaches by calculating the importance of weights \cite{kirkpatrick2017overcoming, zenke2017continual}, to modify the architecture of neural networks \cite{rusu2016progressive, yoon2018lifelong} and to introduce an episodic memory to store and replay the previously learned samples \cite{lopez2017gradient, chaudhry2018efficient}. 

The basic idea of rehearsal-based approaches like Gradient Episodic Memory (GEM) \cite{lopez2017gradient} is to ensure gradient-alignment across tasks such that the losses of the past tasks in episodic memory will not increase. Interestingly, this objective coincides with the implicit objective of gradient-based meta-learning algorithms \cite{nichol2018first, riemer2018learning, gupta2020maml}. Further, meta-learning algorithms show promise to generalize better on future tasks \cite{riemer2018learning, javed2019meta}. Meta Experience Replay (MER) \cite{riemer2018learning} integrates the first order meta-learning algorithm Reptile \cite{nichol2018first} with an experience replay module to reduce interference between old and new tasks. To alleviate the slow training speed of MER, Lookahead-MAML (La-MAML) \cite{gupta2020maml} proposes a more efficient meta-objective for online continual learning. La-MAML then incorporates learnable per-parameter learning rates to further reduce the catastrophic forgetting and achieved state-of-the-art performance. 

However, though La-MAML proposes a more efficient objective, it still requires directly computing the Hessian matrix which is computationally inhibitable for large networks. Also, learning a learning rate for each parameter entails more computational overhead and increases memory usage.

To overcome these difficulties, in this paper, we present a novel efficient gradient-based meta-learning algorithm for online continual learning. The proposed method solves the meta-optimization problem without accessing the Hessian information of the empirical risk. Inspired by regularization-based methods, we compute the parameter importance using the first-order Taylor series and assign the learning rates according to the parameter importance. In this way, no extra trainable parameters will be incorporated so that the computational complexity and memory usage can be reduced. We also impose explicit regularization terms in the inner loss to achieve better performance and apply proximal gradient descent to improve efficiency. Our approach performs competitively on four commonly used benchmark datasets, achieving better or on-par performance against La-MAML and other state-of-the-art approaches in a much shorter training time.

\section{Related Work}

\subsection{Continual Learning}

Existing continual learning approaches are mostly broadly classified into \emph{regularization-based}, \emph{rehearsal-based}, and \emph{dynamic network architecture-based} approaches \cite{parisi2019continual}.

\textbf{Regularization-based methods} penalize major changes by quantifying parameter importance on previous tasks while using a fixed capacity. Parameter importance could be estimated by Fisher information matrix \cite{kirkpatrick2017overcoming}, loss \cite{zenke2017continual} or outputs sensitivity with respect to the parameters \cite{aljundi2018memory} and trainable attention masks \cite{serra2018overcoming}. A number of studies restrain weight updates from Bayesian perspectives \cite{ahn2019uncertainty, ebrahimi2019uncertainty, nguyen2018variational, ritter2018online, titsias2020functional}. Several recently proposed methods also consider forcing weight updates to belong to the null space of the feature covariance \cite{tang2021layerwise, wang2021training}.

\textbf{Rehearsal-based methods} maintain a small episodic memory of previously seen samples for replay \cite{lopez2017gradient, rebuffi2017icarl, chaudhry2018efficient} or train a generative model to produce pseudo-data for past tasks \cite{shin2017continual, kemker2018fearnet, ostapenko2019learning}. Generative models reduce working memory effectively but invoke the complexity of the generative task. In contrast, episodic memory methods are simpler and more effective. Gradient Episodic Memory (GEM) \cite{lopez2017gradient} aligns gradients across tasks to avoid interference with the previous tasks. Averaged-GEM (A-GEM) \cite{chaudhry2018efficient} simplifies GEM by replacing all gradients to one gradient of a sampled batch. Experience Replay (ER) \cite{chaudhry2019continual} considers the online setting and jointly trains the model on the samples from new tasks and episodic memory. A number of methods focus on improving the memory selection process, like MIR \cite{aljundi2019online} that selects most interfered samples for memory rehearsal, HAL \cite{chaudhry2021using} that selects the anchor points of past tasks and interleaves them with new tasks for future training, and GMED \cite{jin2021gradient} that  edits stored examples via gradient updates to create more “challenging” examples for replay. 

\textbf{Dynamic network architectures-based methods} overcome catastrophic forgetting by dynamically allocating task-specific parameters to accommodate new tasks. In \cite{rusu2016progressive, von2020continual, yoon2018lifelong, sarwar2019incremental, xu2018reinforced}, the model is expanded for each new task. Progressive Neural Network (PNN) \cite{rusu2016progressive} leverages prior knowledge via lateral connections to previously learned features. Dynamically Expandable Network (DEN) \cite{yoon2018lifelong} splits or duplicates important neurons on new tasks when expanding the network to reduce such redundancy, whereas \cite{sarwar2019incremental} shares part of the base network. Reinforced Continual Learning (RCL) \cite{xu2018reinforced} searches for the best network architecture for arriving tasks using reinforcement learning. To ensure that the model maintains the compactness, \cite{hung2019compacting} performs wights pruning after training on each task, which highly increases the computational overhead. Dirichlet process mixture models have also been applied to expand a set of networks \cite{jerfel2018reconciling}. Instead of learning the weights of the sub-networks, \cite{mallya2018piggyback, wortsman2020supermasks} find binary masks to assign different subsets of the weights for different tasks. By design, these approaches often result in higher model and time complexities.

\subsection{Meta-Learning for Continual Learning}

Recently, it has been shown that gradient-based meta-learning algorithms integrated with episodic memory outperform many previous approaches in online settings \cite{riemer2018learning, gupta2020maml, javed2019meta}. Meta-Experience Replay (MER) \cite{riemer2018learning} aligns gradients between old and new tasks using samples from an experience replay module. However, the training speed of MER is pretty slow so it's impractical to extend it to real-world scenarios. Online-aware Meta-Learning (OML) \cite{javed2019meta} proposes a meta-objective to learn a sparse representation offline. Lookahead-MAML (La-MAML) \cite{gupta2020maml} introduces a more efficient online objective and incorporates trainable parameter-specific learning rates to reduce the interference. Both La-MAML and MER require the computation of second-order derivatives.

\section{Preliminaries}

\subsection{Continual Learning}

Suppose that a sequence of $T$ tasks $[\tau_1, \tau_2, \dots, \tau_T ]$ is observed sequentially. Each task $\tau_t$ is associated with a dataset $\{ X^t, Y^t \} = \{ (x_m^t, y_m^t)\}_{m=1}^{n_t}$ of $n_t$ example pairs. At any time-step $j$ during online learning, we would like to minimize the loss on all the $t$ tasks seen so far ($\tau_{1:t}$):

\begin{equation}
\begin{aligned}
    \theta^j &= \arg \min_{\theta^j} \sum_{i=1}^t \mathbb{E}_{\tau_i} \big [ \ell_i (\theta^j) \big ] \\
    &= \arg \min_{\theta^j} \mathbb{E}_{\tau_{1:t}} \big [ L_t (\theta^j) \big ]
\end{aligned}
\end{equation}

where $\ell_i$ is the loss on $\tau_i$ using $\theta^j$, the learnt model parameters at time-step $j$. $L_t = \sum_{i=1}^t \ell_i$ is the sum of all task-wise losses for tasks $\tau_{1:t}$. GEM \cite{lopez2017gradient} reformulates this problem as:

\begin{gather}
  \min_{\tilde{g}} \frac{1}{2} \| g - \tilde{g} \|_2^2, \> s.t. \> \langle \tilde{g}, g_p \rangle \geq 0, \> \forall p < t,
\end{gather}

where $g$ and $g_p$ are the gradient vectors computed on the current task and previous tasks $k$ respectively. Such objective can also be treated as maximizing the dot products between gradients of a set of tasks \cite{gupta2020maml}:

\begin{equation}
  \theta^j = \arg \min_{\theta^j} \Bigg ( \sum_{i=1}^t \ell_i (\theta^j) - \alpha \sum_{p,q \leq t} \bigg ( \frac{\partial \ell_p (\theta^j)}{\partial \theta^j} \cdot \frac{\partial \ell_q (\theta^j)}{\partial \theta^j} \bigg ) \Bigg ), \label{eq:gem}
\end{equation}

where $\alpha$ is a trade-off hyper-parameter.

\subsection{Model-Agnostic Meta-Learning}

Model Agnostic Meta-Learning (MAML) \cite{finn2017model} is an gradient-based meta-learning approach aiming to learn meta-parameters that produce good task specific parameters after adaptation. Meta-parameters are learned in the \emph{meta-update} (outer-loop), while task-specific models are learned in the \emph{inner-update} (inner-loop). In every meta-update, its objective at time-step $j$ can be formulated as below:

\begin{equation}
  \underbrace{\min_{\theta_0^j} \mathbb{E}_{\tau_{1:t}} \bigg [ L_{\text{meta}} \Big ( \overbrace{U_k(\theta_0^j)}^{\text{inner-loop}} \Big) \bigg ]}_{\text{outer-loop}} = \min_{\theta_0^j} \mathbb{E}_{\tau_{1:t}} \bigg [ L_{\text{meta}} \Big (\theta_k^j \Big) \bigg ],
\end{equation}

where $\theta_0^j$ is the meta-parameter  at time-step $j$ and $U_k(\theta_0^j) = \theta_k^j$ represents an update function where $\theta_k^j$ is the parameter after $k$ steps of stochastic gradient descent.

\cite{nichol2018first} has proved that MAML and its first-order variation like Reptile approximately optimize for the same objective that gradients are encouraged to align within-task and across-task as well. \cite{riemer2018learning} then showed the equivalence between the objective of GEM (Eq (\ref{eq:gem})) and Reptile. This implies that the procedure to meta-learn an initialization coincides with learning optimal parameters for continual learning.

\subsection{Meta-Learning for Continual Learning}

Although gradient-based meta-learning algorithms implicitly align the gradients, there can still be some interference between the gradients of old tasks $\tau_{1:t-1}$ and new task $\tau_t$. Specifically, when starting training on task $\tau_t$, the gradients are not necessarily aligned with the old ones, since the data of $\tau_{1:t-1}$ is no longer available to us. To ensure meta-updates are conservative with respect to $\tau_{1:t-1}$, MER modifies the Reptile algorithm to integrate it with an experience replay module, which aligns gradients between old and new tasks during meta-updates. However, MER's algorithm is prohibitively slow in training speed. 

La-MAML \cite{gupta2020maml} then introduces two mechanisms as follows. (1) Optimizing for an alternative objective:

\begin{equation}
  \label{eq:la-maml-objective}
  \min_{\theta_0^j} \sum_{\mathcal{S}_k^j \sim D_t} \bigg [ L_t \Big (U_k(\theta_0^j, \mathcal{S}_k^j) \Big) \bigg ],
\end{equation}

where $\mathcal{S}_k^j = \big \{ x^{t}_{j+l}, x^{t}_{j+l} \big\}_{l=1}^k$ is a random stream of length $k$ sampled from $\{X^t, Y^t\}$ at time-step $j$. The meta-loss $L_t = \sum_{i=1}^t \ell_i$ is evaluated on $\theta_k^j = U_k (\theta_0^j, S_k^j)$. Eq (\ref{eq:la-maml-objective})'s objective only uses one data point from $\mathcal{S}_k^j$ for one inner-update instead of using the complete batch of data $\mathcal{S}_k^j$ for all inner-updates, which significantly improves the computational and memory efficiency. It appears to be the relationship between SGD and batch-GD. La-MAML then proves that this mechanism coincides with AGEM's objective, i.e., aligning the gradients of $\tau_t$ and the \emph{average gradient} of $\tau_{1:t}$ instead of aligning all the pair-wise gradients between tasks $\tau_{1:t}$. (2) Incorporating learnable per-parameter learning rates to mitigate catastrophic forgetting. Learning rates are modulated during each meta-update and used for both inner-update and meta-update.

La-MAML shows impressive performance and drastic speedup when compared to other meta-learning based methods. However, it still relies on directly computing the Hessian matrix during meta-update, which incurs a lot of computational overhead. Also, learning a separate learning rate for each parameter requires increased computational effort and larger memory usage.

\section{Proposed Method}

In this section, based on La-MAML's scheme and inspired by some regularization-based methods, we propose Efficient Meta-Learning for Continual Learning (EMCL), a novel approach that neither requires computing the Hessian matrix nor incorporates any other learnable parameters.

\subsection{Inner-Update with Explicit Regularization}

Different from La-MAML where the same meta-update learning rate is also used for the inner-update, we empirically found that explicitly adding regularization terms to the inner loss function achieves better performance:

\begin{equation}
  \begin{aligned}
  \theta_k^j &= \arg \min_{\theta_k^j} \ell_i (\theta_k^j) \\
    &= \arg \min_{\theta_k^j} \left \{ \mathcal{L}(\theta_k^j) + \frac{\lambda}{2} \sum_m h_m^j \big \| \theta_{k, m}^j - \theta_{0, m}^j \big \|_2^2 \right \} \\
    &= \arg \min_{\theta_k^j} \left \{ \mathcal{L}(\theta_k^j) + \frac{\lambda}{2} \left \| \mathbf{H}^j (\theta_k^j - \theta_0^j) \right \|_2^2 \right \}.
  \end{aligned} \label{eq:explicit-reg-inner}
\end{equation}

$\mathcal{L}$ is the empirical risk function. $\mathbf{H}^j$ is a diagonal matrix with $\mathbf{H}^j_{m,m} = \sqrt{h^j_m}$, where $h_m^j$ is the moving average of the importance of the $m$-th parameter $\Omega_m^j$ at time-step $j$:

\begin{equation}
h_m^j = \eta h_m^{j-1} + (1 - \eta) \Omega_m^j.
\end{equation}

$\eta$ is the decay rate and $r$ is a scale factor. The rationale of using the moving average is that even a small change in important parameters may drastically degrade the performance of previous tasks. We will explain the way of computing $\Omega_m^j$ later in Section \ref{sec:parameter-importance}. This regularization term can be viewed as an importance-weighted $\ell_2$ norm. On the one hand, the regularization term in Eq (\ref{eq:explicit-reg-inner}) tries to reduce the change in important weights. On the other hand, it also encourages $\theta_k^j$ to remain close to $\theta_0^j$, thereby retaining a strong dependence on the initial parameters to avoid over-fitting and vanishing gradients \cite{rajeswaran2019meta}. This point is important especially when multiple inner-loop optimization steps are required.

\subsection{Closed-form Meta-Update with Adaptive Learning Rate}

 During meta-update, we scale the learning rate for each parameter inversely proportional to the moving average of its importance. Let $\alpha_m^j$ be the learning rate of the $m$-th meta-parameter at time-step $j$, we have:

\begin{equation}
    \alpha_m^j \leftarrow \frac{r}{h_m^j} \alpha_m^{j-1}, \label{eq:adaptive_meta_lr}
\end{equation}

where $r$ is a scale factor. In this way, changes in important parameters can be reduced, while less important parameters allow having larger step sizes in future tasks.

Supposing that $\theta_k$ is the unique minimizer of Eq (\ref{eq:explicit-reg-inner}), we know that the gradient of the inner-loss should be zero:

\begin{equation}
  \nabla L(\theta_k) + \lambda \mathbf{H}^2 (\theta_k - \theta_0) = 0.
\end{equation}

Here we omit the superscript $j$ for convience. Then the gradient of the meta-loss can be expressed in closed-form as:

\begin{equation}
  \label{eq:outer-update}
  \begin{aligned}
    \nabla_{\theta_0} L_t(\theta_k) &= \left (\frac{\partial \theta_k}{\partial \theta_0} \right )^{\top} \nabla L (\theta_k) \\ & \qquad \quad + \lambda \left( \left( \frac{\partial \theta_k}{\partial \theta_0} \right)^{\top} - I \right) \mathbf{H}^2 (\theta_k - \theta_0) \\
      &= \frac{\partial \theta_k}{\partial \theta_0} \left (\nabla L (\theta_k) + \lambda \mathbf{H}^2 (\theta_k - \theta_0) \right ) - \lambda \mathbf{H}^2 (\theta_k - \theta_0) \\
      &= \lambda \mathbf{H}^2 (\theta_0 - \theta_k).
  \end{aligned}
\end{equation}

This algorithm is a first-order method without the need of computing the Hessian matrix. However, unlike other first-order methods like Reptile \cite{nichol2018first} and FOMAML \cite{finn2017model}, it can make use of higher-order information of the inner loss beyond gradients to search the optimal hypothesis around $\theta_0$ as well as to offer the theoretical guarantees of convergence and generalization. The proof can be easily extended from \cite{zhou2019efficient}'s work.

\subsection{Parameter Importance Estimation via Taylor Expansion} \label{sec:parameter-importance}

We update the estimation of importance for each parameter in every meta-update. The importance of the $m$-th meta-parameter can be quantified by the impact on the total loss after zeroing it out:

\begin{equation}
  \Omega_m = \bigg | L_t(\theta_0) - L_t \left (\theta_0 \big |_{\theta_{0, m} = 0} \right ) \bigg |.
\end{equation}

Here $\theta_0^j \big |_{\theta_{0, m}^j = 0}$ indicates the zeroing out operation. In essence, zeroing out is similar to \emph{leave-one-out} in classification. It is computationally intensive to computing $\Omega_{k}$ for each parameter in every meta-update, so we approximate it using first-order Taylor expansion. The Taylor series of loss function $L(\theta_0^j)$ at $\theta_0 \big |_{\theta_{0, m} = 0}$ is:

\begin{equation}
  \label{eq:taylor_expansion}
  L_t \left (\theta_0 \big |_{\theta_{0, m} = 0} \right ) = L_t(\theta_0) + \frac{\partial L_t(\theta_0)}{\partial \theta_{0, m}} \big ( \theta_{0, m} - 0 \big ) + o(\theta_{0, m}),
\end{equation}

where $o(\theta_{0, m})$ represents the terms of higher orders. We omit the higher-order terms in Eq (\ref{eq:taylor_expansion}) to obtain an approximate parameter importance:

\begin{equation}
  \label{eq:taylor_importance}
  \Omega_m = \Bigg | L_t \left (\theta_0 \big |_{\theta_{0, m} = 0} \right ) - L_t (\theta_0) \Bigg | \approx \left | \frac{\partial L_t (\theta_0)}{\partial \theta_{0, m}} \theta_{0, m} \right |.
\end{equation}

Computing Eq (\ref{eq:taylor_importance}) incurs little computational overhead, since the gradient $\partial L_t (\theta_0) / \partial \theta_{0, m}$ is already available after inner-update. It should be noted that, Eq (\ref{eq:taylor_importance}) computes the weight importance using the absolute value of the product of the gradient with the \emph{parameter}, which is different from that used by SI \cite{zenke2017continual}, where the importance is calculated using the product of the gradient with the \emph{parameter update}.

\subsection{Inner-Update with Proximal Gradient Descent}

While Eq (\ref{eq:explicit-reg-inner}) can be minimized via applying normal gradient descent methods, we employ the Proximal Gradient Descent (PGD) method. We first denote the proximal operator of a function $f(\cdot)$ with a scalar parameter $\gamma > 0$ as:

\begin{equation}
  \text{prox}_{\gamma f} (v) = \arg \min_{x} \left ( f(x) + \frac{1}{2 \gamma} \| x - v \|^2_2 \right ),
\end{equation}

where $x \in \mathbb{R}^n, v \in \mathbb{R}^n$ are two $n$ dimensional vectors. In this case, for $\theta_{k,m}$, we have:

\begin{equation} \label{eq:pgd-f}
f(\theta_{k,m}) = \frac{\lambda}{2} h_m \| \theta_{k, m} - \theta_{0, m} \big \|_2^2. 
\end{equation}

Proximal operators can be interpreted as modified gradient steps:

\begin{gather}
  \hat{\theta}_\kappa = \theta_{\kappa-1} - \gamma \nabla L (\theta_{\kappa-1}), \label{eq:pgd-1} \\
  \theta_\kappa = \text{prox}_{\gamma f} (\hat{\theta}_\kappa), \label{eq:pgd-2}
\end{gather}

for $\kappa = \{ 1, \dots, k \}$. We then introduce the following lemma: 

\vspace{8pt}

\begin{lemma}
  \label{lemma:pgd}
  For $f(x) = c \| x - x_0 \|_2^2$ with $c > 0$ and any fixed vector $x_0$, we have:
\end{lemma}

\begin{equation}
  \text{prox}_{\gamma f} (v) = \frac{v + 2 \gamma c x_0}{2 \gamma c + 1}.
\end{equation}

\begin{proof}
  $\text{prox}_{\gamma f}$ minimizes the function:
\end{proof}

\begin{equation}
  l(x) = \| c (x - x_0) \|_2^2 + \frac{1}{2 \gamma} \| x - v \|^2_2 .
\end{equation}

We denote $x^*$ as the minimizer of $l(x)$ and we know that $\nabla l(x^*) = 0$ since $l(x)$ is convex. By differentiation, we get:

\begin{equation}
  \begin{aligned}
    \nabla l(x) = 2 c (&x^* - x_0) + \frac{1}{\gamma} (x^* - v) = 0, \\
    \Rightarrow x^* &= \frac{v + 2 \gamma c x_0}{2 \gamma c + 1}.
  \end{aligned}
\end{equation}

By Lemma \ref{lemma:pgd}, we have the following closed-form proximal gradient update rule for Eq (\ref{eq:pgd-f}):

\begin{equation}
  \label{eq:pgd-update-rule}
  \theta_{\kappa, m} = \frac{\hat{\theta}_{\kappa, m} + \gamma \lambda h_m \theta_{0,m}}{\gamma \lambda h_m + 1}.
\end{equation}

Eq (\ref{eq:pgd-update-rule}) is expected to be more efficient in computation than directly applying vanilla gradient decent methods since it avoids computing gradients for the quadratic regularization term in Eq (\ref{eq:explicit-reg-inner}). We also find it achieves higher accuracy (see ablation results in Table \ref{table:result-ablation}).

Finally, we summarize our method EMCL in Algorithm \ref{alg:our_method}.

\begin{algorithm} \caption{EMCL (LR: Learning Rate)} \label{alg:our_method}
  \begin{algorithmic}[1]
  \Require Network weights $\theta_0^0$, inner objective $\ell$, meta objective $L$, inner LR $\beta$, meta LRs $\alpha$, parameter importance decay rate $\eta$, LR scale factor $r$, inner regularization $\lambda$
  \State \textbf{initialize:} $j \gets 0$
  \For{$t := 1 \to T$}
    \For{batch $b$ in $\{X^t, Y^t\}$}
      \State $k \gets \text{sizeof}(b)$
      \For{$\kappa = 1 \to k$}
        \State $\theta_\kappa^j \gets \theta_{\kappa-1}^j - \beta \nabla_{\theta_\kappa^j} \ell_t(\theta_\kappa^j, b[\kappa], \lambda)$ \algorithmiccomment{optimize for Eq (\ref{eq:explicit-reg-inner}) via (\ref{eq:pgd-1}), (\ref{eq:pgd-2}) and (\ref{eq:pgd-update-rule})}
      \EndFor
      \State $\alpha^{j+1} \gets \text{UpdateMetaLR}(\alpha^j, \eta, r)$ \algorithmiccomment{modulate meta LR accroding to Eq (\ref{eq:adaptive_meta_lr})}
      \State $\theta_0^{j+1} \gets \theta_0^j - \alpha^j \nabla_{\theta_0^j} L_t (\theta_k^j, b)$ \algorithmiccomment{meta-update via Eq (\ref{eq:outer-update})}
      \State $j \gets j + 1$
    \EndFor
\EndFor
\end{algorithmic}
\end{algorithm}

\begin{table*}[!htbp]
  \caption{Statistics of the benchmark datasets}
  \centering
  \label{teble:dataset}
    
  \begin{tabular}{lcccc}
    \toprule
      & MNIST Perm. & Many Perm. & CIFAR100 & miniImageNet \\
    \midrule
    Input size (color and image pixels) & 1$\times$28$\times$28 & 1$\times$28$\times$28 & 3$\times$32$\times$32 & 3$\times$84$\times$84 \\
    Num of tasks & 20 & 100 & 20 & 20 \\
    Num of samples per task & 1000 & 200 & 2500 & 2500 \\
    Num of classes per task & 10 & 10 & 5 & 5 \\
    \bottomrule
  \end{tabular}
\end{table*}

\begin{table*}[!htbp]
  \centering
  \caption{List of hyper-parameters for all approaches}
  \label{table:hyper}
  \begin{tabular}{lccccc}
    \toprule
    Method & Hyper Parameter & MNIST Perm & Many Perm & CIFAR100 & miniImageNet \\
    \midrule
    \multirow{2}*{EWC} & lr & - & - & 0.03 & 0.03 \\
      & regularization & - & - & 100 & 100 \\
    \midrule
    \multirow{2}*{GEM} & lr & - & - & 0.03 & 0.03 \\
      & memory size & - & - & 200 & 200 \\
    \midrule
    \multirow{2}*{A-GEM} & lr & - & - & 0.03 & 0.03 \\
      & memory size & - & - & 200 & 200 \\
    \midrule
    \multirow{3}*{La-MAML} & $\alpha_0$: initial lr & 0.3 & 0.1 & 0.1 & 0.1 \\
      & $\eta$: lr for $\alpha_0$ & 0.15 & 0.1 & 0.3 & 0.3 \\
      & memory size & 200 & 500 & 200 & 200 \\
    \midrule
    \multirow{4}*{EMCL} & $\alpha_0$: initial meta lr & 0.3 & 0.15 & 0.1 & 0.1 \\
      & $\beta$: inner lr & 0.15 & 0.03 & 0.075 & 0.075 \\
      & $\lambda$: regularization & 10 & 10 & 50 & 50 \\
      & $\gamma$: scalar for pgd & 0.3 & 0.1 & 0.1 & 0.1 \\
    \bottomrule
  \end{tabular}
\end{table*}

\section{Experimental Setup} \label{sec:experimental_setup}

In this section, we perform experimental evaluations of our method compared with existing state-of-the-art methods for continual learning on four benchmark datasets.

\textbf{Datasets}: We perform experiments on four benchmark datasets: 

\begin{itemize}
  \item \textbf{MNIST Permutations} \cite{kirkpatrick2017overcoming} is a variant of MNIST, where each task is a random permutation of the original MNIST pixels. As such, the input
  distribution of each task is unrelated. It has 20 tasks, each with
  1000 samples from 10 different classes.
  \item \textbf{Many Permutations} \cite{riemer2018learning} is a variant of MNIST Permutations that has 5 times more tasks (100 tasks) and 5 times fewer samples (200 samples) per task.
  \item \textbf{Split CIFAR100} \cite{zenke2017continual} splits the original CIFAR-100 dataset \cite{krizhevsky2009learning} into 20 disjoint tasks, where each consists of 2,500 samples from 5 classes which are not included in other tasks.
  \item \textbf{Split miniImageNet} \cite{vinyals2016matching} is constructed by splitting 100 classes of miniImageNet into 20 sequential tasks, where each task has 2,500 samples from 5 classes.
\end{itemize}

We also summarize the statistics of the benchmark datasets we used in Table \ref{teble:dataset}.

\begin{table*}
  \caption{ACC (\%) (larger is better), BWT (\%) (larger is better) and their standard deviation on MNIST Permutations, Many Permutations, Split CIFAR100, and Split miniImageNet. $\dagger$ denotes the result reported by \cite{gupta2020maml}. Other results are reproduced by us, where each of them is run with 5 random seeds. Due to the slow training speed of MER, we do not report its performance on CIFAR100 and miniImageNet.}
  \label{table:results}
  \centering
  \renewcommand\arraystretch{1.5}{
    \resizebox{\textwidth}{!}{
      \begin{tabular}{ccccccccc}
        \toprule
        \multirow{2}*{Method} & \multicolumn{2}{c}{MNIST Perm.} & \multicolumn{2}{c}{Many Perm.} & \multicolumn{2}{c}{CIFAR100} & \multicolumn{2}{c}{miniImageNet} \\ 
        \cline{2-9}
          & ACC   & BWT   & ACC   & BWT & ACC   & BWT   & ACC   & BWT   \\
        \midrule
        EWC & 62.32 $\pm$ 1.34 $\dagger$ & -13.32 $\pm$ 2.24 $\dagger$ & 33.46 $\pm$ 0.46 $\dagger$ & -17.84 $\pm$ 1.15 $\dagger$ & 39.60 $\pm$ 1.11 & -23.53 $\pm$ 1.19 & 34.34 $\pm$ 2.06 & -28.17 $\pm$ 1.49 \\
        GEM & 55.42 $\pm$ 1.10 $\dagger$ & -24.42 $\pm$ 1.10 $\dagger$ & 32.14 $\pm$ 0.50 $\dagger$ & -23.52 $\pm$ 0.87 $\dagger$ & 43.41 $\pm$ 2.09 & -20.76 $\pm$ 1.31 & 37.02 $\pm$ 1.91 & -25.29 $\pm$ 2.10 \\
        A-GEM & 56.04 $\pm$ 2.36 & -24.05 $\pm$ 2.47 & 29.98 $\pm$ 1.84 & -27.23 $\pm$ 1.79 & 43.87 $\pm$ 2.61 & -23.38 $\pm$ 1.52 & 36.37 $\pm$ 1.56 & -25.11 $\pm$ 2.92 \\
        MER & 73.46 $\pm$ 0.45 $\dagger$ & -9.96 $\pm$ 0.45 $\dagger$ & 47.40 $\pm$ 0.35 $\dagger$ & -17.78 $\pm$ 0.39 $\dagger$ & - & - & - & - \\
        La-MAML & \textbf{73.92 $\pm$ 1.05} & -7.91 $\pm$ 0.87 & 47.69 $\pm$ 0.41 & -13.24 $\pm$ 0.95 & 61.23 $\pm$ 0.94 & -19.84 $\pm$ 2.20  & 45.29 $\pm$ 1.76 & -18.57 $\pm$ 2.94  \\
        \midrule
        \textbf{EMCL} & 73.61 $\pm$ 1.12 & -10.25 $\pm$ 0.73 & \textbf{48.12 $\pm$ 1.48} & -14.09 $\pm$ 0.74 & \textbf{61.95 $\pm$ 1.20} & -16.48 $\pm$ 1.96 & \textbf{46.52 $\pm$ 0.83} & -17.45 $\pm$ 2.38 \\
        \bottomrule
      \end{tabular}
    }
  }
\end{table*}

\begin{figure*}
  \centering
  \label{fig:acc-evolution}
  \includegraphics[width=\textwidth]{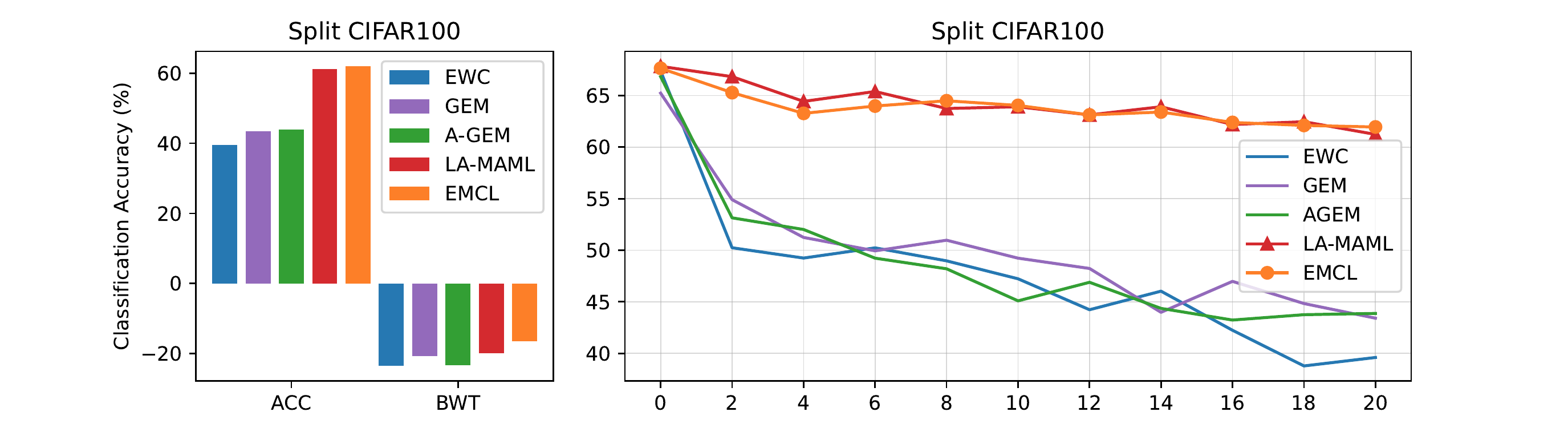}
  \caption{Left: ACC and BWT for all appraoches on CIFAR100. Right: evolution of the average test accuracy as more tasks are learned.}
\end{figure*}

\begin{figure}
  \centering
  \label{fig:training-time}
  \includegraphics[width=0.35\textwidth]{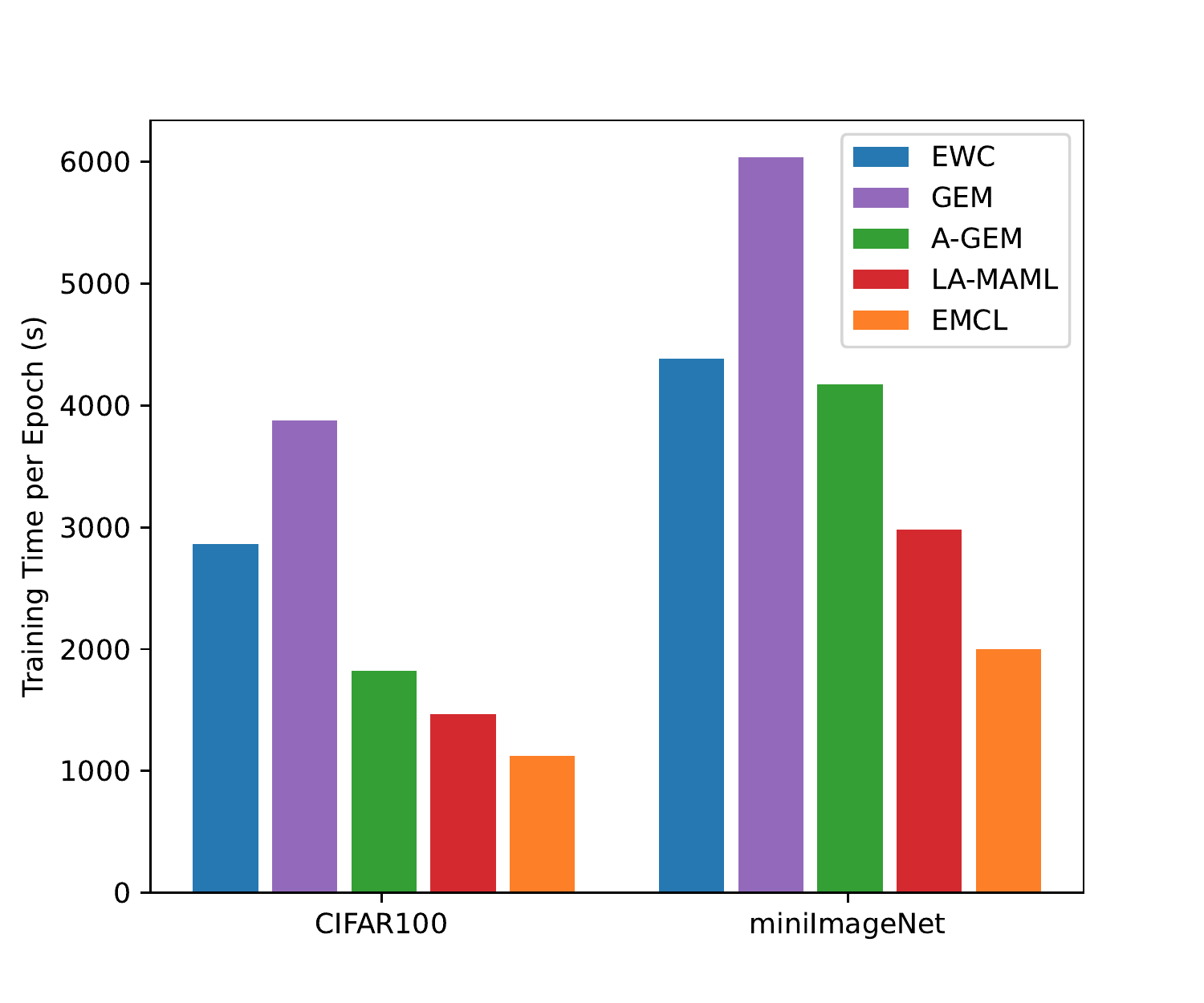}
  \caption{Training time for all algorithms on Split CIFAR100 and miniImageNet.}
\end{figure}

\textbf{Training Details}: Unlike most prior works that train each task in multiple epochs, we consider a more challenging setup following \cite{gupta2020maml}: \emph{single-pass} setting formalized by \cite{chaudhry2018efficient} assumes data for every task to be processed for only one epoch. After that, data samples are not accessible anymore unless they were added to a replay buffer. We use this setting for all our experiments.

\emph{Single-head} setting is used for experiments on MNIST Permutations and Many Permutations, where all tasks share the final classifier layer and inference is performed without task identity. For other experiments, we use \emph{multi-head} setting, where
each task has a separate classifier.

For methods that require storing the past samples (GEM, A-GEM, MER, and La-MAML), we use a replay buffer of size 200, 500, 200, and 200 for MNIST Permutations, Many Permutations, Split CIFAR100, and Split miniImageNet, respectively. We carry out hyper-parameter tuning by performing a grid-search for hyper-parameters related to the learning-rate. See Table \ref{table:hyper} for more details of hyper-parameters.

\textbf{Model Architecture}: We use a fully-connected network with two hidden layers of 100 ReLU units each for MNIST Permutations and Many Permutations. The architecture is the same as in MER \cite{riemer2018learning} and La-MAML \cite{gupta2020maml} to compare the results directly. For Split CIFAR100 and Split miniImageNet, we use a reduced ResNet18 \cite{he2016deep} architecture with three times fewer feature maps across all layers by following \cite{lopez2017gradient}. 

For a given dataset, all networks use the same architecture and are optimized via SGD optimizer with a batch size of 10 samples from an online stream \cite{lopez2017gradient}.

\textbf{Baselines}: We compare our method against the following baselines:

\begin{itemize}
  \item \textbf{EWC}: Elastic Weight Consolidation \cite{kirkpatrick2017overcoming} is an algorithm where the loss is regularized by Fisher Information to avoid catastrophic forgetting.
  \item \textbf{GEM}\footnote{The code of GEM and EWC is adopted from\\ https://github.com/facebookresearch/GradientEpisodicMemory}: Gradient Episodic Memory \cite{lopez2017gradient} does constrained optimization by solving a quadratic program on the gradients of new and replay samples, so that they do not interfere with past memories.
  \item \textbf{A-GEM}: Averaged Gradient Episodic Memory \cite{chaudhry2018efficient} improves GEM by replacing all gradients to one gradient of a sampled batch such that the average episodic memory loss will not increase.
  \item \textbf{MER}: Meta Experience Replay \cite{riemer2018learning} samples i.i.d data from a replay memory to meta-learn model parameters that show increased gradient alignment between old and current samples.
  \item \textbf{La-MAML}\footnote{The code of A-GEM and La-MAML is adopted from\\ https://github.com/montrealrobotics/La-MAML}: Look-ahead MAML \cite{gupta2020maml} uses learnable per-parameter learning rates to mitigate catastrophic forgetting aided by a episodic memory.
\end{itemize}

\textbf{Performance Metrics}: We report the following metrics by evaluating the model on the test set:

\begin{itemize}
  \item \textbf{Accuracy (ACC)}: Average test accuracy of all $T$ tasks after the whole learning is finished: 
  \begin{equation}
    \text{ACC} = \frac{1}{T} \sum_{t=1}^T \text{ACC}_{t, T}
  \end{equation}
  where $\text{ACC}_{t, T}$ is the accuracy of task $t$, after finishing the training process on task $T$.
  \item \textbf{Backward Transfer (BWT)}: The average influence of new learning on past knowledge. For instance,
  negative BWT indicates catastrophic forgetting. It is formally defined as:
  \begin{equation}
    \text{BWT} = \frac{1}{T-1} \sum_{t=1}^{T-1} \text{ACC}_{t, T} - \text{ACC}_{t, t}
  \end{equation}
\end{itemize}

\section{Results and Discussions} \label{sec:result}

\textbf{Performance:} Table \ref{table:results} shows the overall experimental results. We also visualize the ACC, BWT, and evolution of the average test accuracy as a function of the number of tasks for CIFAR100 in Figure \textcolor{red}{1}.

In every setting, EMCL outperforms the commonly used baselines like EWC \cite{lopez2017gradient}, GEM \cite{kirkpatrick2017overcoming} and A-GEM \cite{chaudhry2018efficient} significantly, especially on complex datasets such as Split CIFAR100 and Split miniImageNet. Even when compared to strong baselines like MER \cite{riemer2018learning} and La-MAML \cite{gupta2020maml}, our approach is still able to achieve better or comparable performance while taking shorter training time (Figure \textcolor{red}{2}). It is also worth emphasizing that the number of trainable parameters in La-MAML is twice as much as ours.

Although \cite{chaudhry2018efficient} have mentioned that EWC and similar regularization-based methods perform substantially worse than rehearsal-based methods like GEM and A-GEM on the single-pass setting, they achieve similar accuracy in our experiments. This is possibly due to the size of replay buffers for rehearsal-based methods in our experiments (1-2 samples per class) being much smaller than those in theirs (13-25 samples per class). Such results demonstrate that rehearsal-based methods are not very effective in a low-resource regime. Even so, rehearsal-based methods still require an external episodic memory to store images from past tasks. While EMCL only needs a much smaller memory for storing the latest weighted moving averages of parameter importance and per-parameter learning rates. 

\textbf{Training Time}: Figure \textcolor{red}{2} shows the training time after learning all tasks per epoch for different algorithms. The training time is measured on a single NVIDIA GeForce RTX 2080 GPU, including time spent for memory management for GEM, A-GEM, and La-MAML and weight importance calculation for EWC and EMCL. We see EMCL takes a much shorter training time than EWC, GEM, and A-GEM while achieving much higher accuracy. EMCL is also faster than La-MAML because it doesn't need to compute second-order derivatives and learn per-parameter learning rates.

\begin{table}
  \centering
  \caption{Ablation study on Split miniImageNet}
  \label{table:result-ablation}
  \begin{tabular}{lcccc}
    \toprule
    Inner LR modulation & $\times$ & $\times$ & $\checkmark$ \\
    Inner regularization & $\checkmark$ & $\checkmark$ & $\times$ \\
    PGD & $\checkmark$ & $\times$ & $\times$ \\
    \midrule
    ACC (\%) & \textbf{46.52} & 46.17 & 45.04 \\
    \bottomrule
  \end{tabular}
\end{table}

\begin{figure}
  \centering
  \label{ablation-acc-vs-training-time}
  \includegraphics[width=0.43\textwidth]{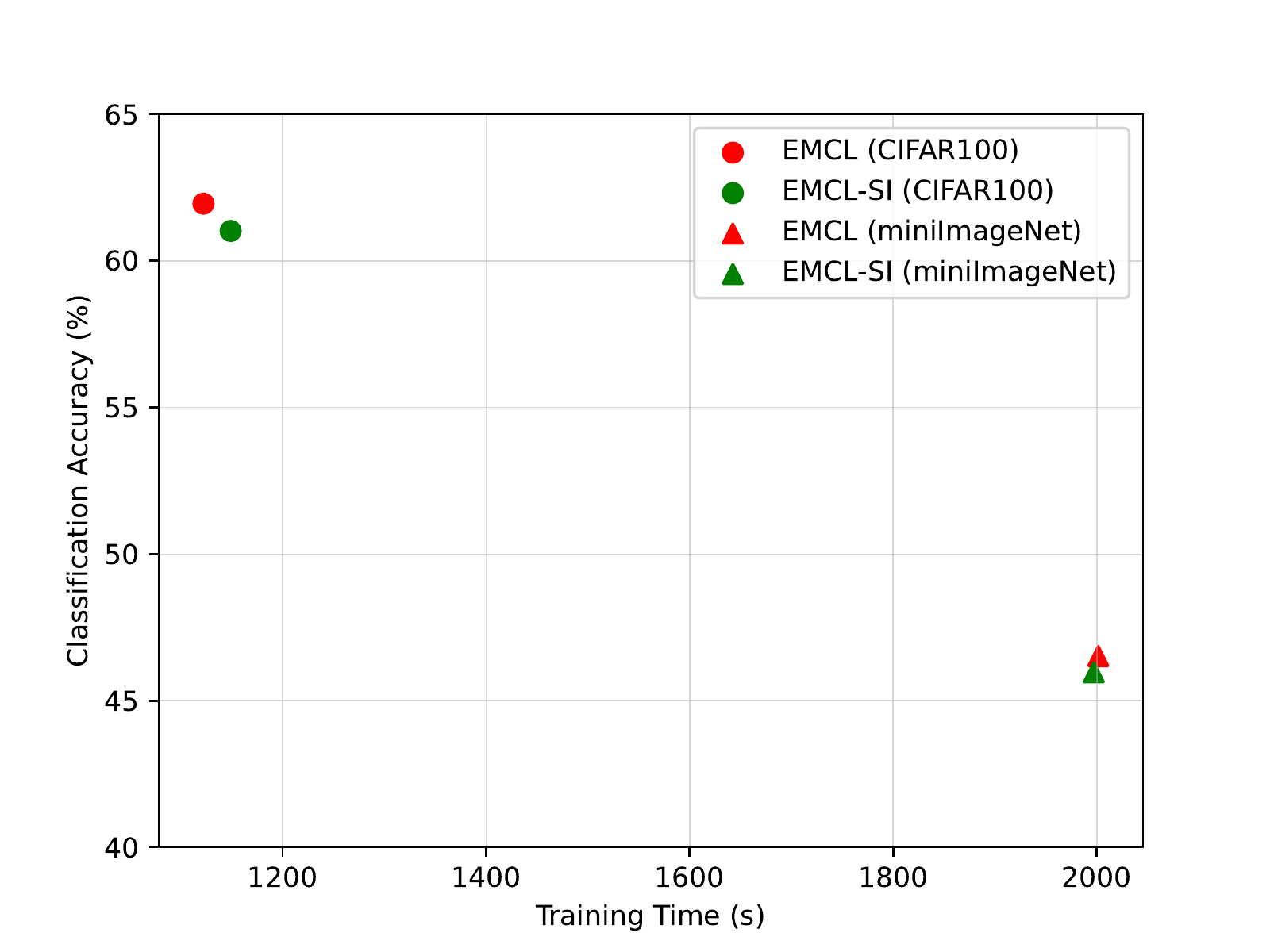}
  \caption{Accuracy vs training time comparison on Split CIFAR100 and miniImageNet for EMCL and EMCL-SI.}
\end{figure}

\textbf{Ablation Study}: We also perform an ablation study and the results are reported in Table \ref{table:result-ablation}. \emph{Inner LR modulation} denotes using the same meta learning rates for the inner loop without adding explicit regularization terms. In this setting, the method performs 1.48\% worse, indicating adding explicit regularization terms to inner loss is more effective in alleviating catastrophic forgetting. We also find that removing proximal gradient decent results in a drop in accuracy. This is possibly due to proximal gradient descent can find an analytic solution for part of the inner objective directly.

We then compare our parameter importance estimation method based on first-order Taylor series to that used by SI \cite{zenke2017continual} where parameter importance is calculated using the product of the gradient and the \emph{parameter update}. The results are summarized in Figure \textcolor{red}{3}. We see that the training speed of EMCL-SI is similar to EMCL but it performs worse slightly.

\section{Conclusion}

In this paper, we propose a novel efficient meta-learning algorithm for continual learning problems. Based on parameter importance estimated using the Taylor series, we modulate the meta-update learning rates and add explicit regularization terms to the inner loss to alleviate catastrophic forgetting. Our method expresses the gradient of meta-updated in closed-form to avoid accessing the Hessian information. We also use proximal gradient descent to solve the inner objective easier and improve the computational efficiency. Experiments on diverse benchmark datasets with different network architectures against strong baselines demonstrate the effectiveness of our approach in achieving high performance in a much shorter time.

\bibliographystyle{IEEEtran}
\bibliography{ref}

\vspace{12pt}

\end{document}